\documentclass[11pt,a4paper]{article}
\pdfoutput=1
\usepackage{emnlp2021}
\usepackage{times}
\usepackage{latexsym}
\usepackage{tabularx}
\usepackage{booktabs}
\usepackage{comment}
\usepackage{multirow}
\usepackage{url}
\usepackage{graphicx}

\graphicspath{{images/}}

\usepackage{pifont} 

\usepackage{microtype}

\newcommand{\cmark}{\ding{51}}

\title{SeqScore: Addressing Barriers to Reproducible Named Entity Recognition Evaluation}

\author{Chester Palen-Michel \and Nolan Holley \and Constantine Lignos \\
  \texttt{\{cpalenmichel,lignos\}@brandeis.edu} \\
  \texttt{nrh2@williams.edu}\\
  Michtom School of Computer Science \\
  Brandeis University
}

\date{}

\begin{document}
\maketitle

\begin{abstract}
To address a looming crisis of unreproducible evaluation for named entity recognition, we propose guidelines and introduce SeqScore, a software package to improve reproducibility.
The guidelines we propose are extremely simple and center around transparency regarding how chunks are encoded and scored.
We demonstrate that despite the apparent simplicity of NER evaluation, unreported differences in the scoring procedure can result in changes to scores that are both of noticeable magnitude and statistically significant.
We describe SeqScore, which addresses many of the issues that cause replication failures.
\end{abstract}

\section{Introduction}

There are many complex tasks in natural language processing (NLP) where current evaluation standards are based around evolving metrics designed to correlate well with human judgments, some complex and some simple.
For example, every year sees the introduction and careful evaluation of new metrics for machine translation, summarization, and natural language generation.

However, named entity recognition (NER) and other chunk extraction tasks have largely been evaluated the same way since the CoNLL shared tasks of the early 2000s \citep{tjong-kim-sang-buchholz-2000-introduction,tjong-kim-sang-2002-introduction,tjong-kim-sang-de-meulder-2003-introduction}.
Following the CoNLL chunking and NER shared tasks, a true positive prediction typically requires exact matches\footnote{While there have been efforts to promote partial matching \citep{chinchor-1998-overview,segura-bedmar-etal-2013-semeval} and focusing on rarer entities \citep{derczynski-etal-2017-results}, micro-averaaged exact match F1 is still the most common metric in use for NER.} in span (the tokens or characters in a chunk) and the type assigned to the chunk (e.g., person).

With such a simple metric, it would seem that performing exact match NER evaluation would be trivially simple.
Precision, recall, and F1 are easy to compute; all that is required is to count true positives, false positives, and false negatives.
But when it comes to evaluation, challenges emerge in how evaluation is actually implemented.
In the case of NER, as we will demonstrate these challenges emerge in the process of converting token-level annotations and system predictions into spans.

We do not think it is sufficient to point out these issues without attempting to provide a solution.
Inspired by successful efforts in the machine translation community to address similar issues \citep{post-2018-call}, we began developing a toolkit and set of practices in summer 2020 to improve the replicability of experiments for NER.
Our toolkit, SeqScore, provides researchers the necessary tools to score, validate, and examine both system outputs and annotation.
SeqScore is open source and has been publicly released.\footnote{\url{https://github.com/bltlab/seqscore}}

This paper provides clear, easy-to-follow guidelines that facilitate reproducibility for NER (and other chunking task) experiments, explains them, provides a toolkit for easily following them, and then presents experiments using SeqScore that shows the impact of following them.
The contribution of this paper is that it introduces and justifies guidelines for NER experiment reproducibility and provides a toolkit that makes them easy to follow.

\section{Guidelines for reproducibility}

We propose that in order to have sound and reproducible NER evaluation, the following guidelines should be followed:

\begin{enumerate}
    \item \label{rule:report-encoding} Report what chunk encoding scheme was used (e.g. BIO).
    \item \label{rule:external-script} Use an external scorer---not one internal to the system---and report which scorer was used.
    \item \label{rule:repair} Be explicit regarding what form of invalid label sequence repair was used.
    \item \label{rule:gold} Only score against a gold standard that faithfully follows the chunk encoding scheme (e.g. BIO) in use.
    \item \label{rule:stats} Use good statistical practices when reporting results.
\end{enumerate}

Many of these will seem like obvious ideas or practices that should be taken as a given.
However, we have found almost no papers provide enough information to determine whether they are compliant with all of these guidelines specifically, very few papers report what scorer was used to produce the reported scores, and many that provide accompanying code do not include any evaluation code. 

We examined several papers with state of the art NER results on the CoNLL 2003 dataset considering guidelines \ref{rule:report-encoding}, \ref{rule:external-script}, and \ref{rule:repair}.  
Of these papers \citet{liu-etal-2019-gcdt} follow \ref{rule:report-encoding}, \ref{rule:external-script}, and \ref{rule:repair}.
\citet{yamada-etal-2020-luke} explicitly follows guidelines \ref{rule:external-script} and \ref{rule:repair}. 
\citet{luoma-pyysalo-2020-exploring} met guideline \ref{rule:report-encoding}.
\citet{akbik-etal-2019-pooled} give details of their scoring decision for a previous paper, \citet{akbik-etal-2018-contextual}, mentioning they fixed an prior error in scoring, but do not explicitly detail how they fixed their scoring procedure for the baseline in \citet{akbik-etal-2019-pooled}.
All other papers we surveyed did not explicitly satisfy guidelines \ref{rule:report-encoding}, \ref{rule:external-script}, and \ref{rule:repair} \citep{wang-etal-2020-automated,wang-etal-2021-improving,shahzad2021inferner,baevski-etal-2019-cloze,yu-etal-2020-named,jiang-etal-2019-improved,li-etal-2020-unified,devlin-etal-2019-bert}.

As an example of a common departure from these practices, many papers that perform NER experiments publish the scores produced by NCRF++ \citep{yang-etal-2018-design}.
As previously detailed by \citet{lignos-kamyab-2020-build}, NCRF++ uses an internal scorer with an undocumented label sequence repair method, so reporting any numbers from it would be contrary to guidelines \ref{rule:external-script} and \ref{rule:repair}.
As \citeauthor{lignos-kamyab-2020-build} demonstrated, on a specific subset of models that produce a high number of invalid transitions, that scorer produces F1 scores approximately half a point higher than the most commonly-used external scorer. 

Guideline \ref{rule:gold}, which requires that the annotation precisely follow the chunk encoding scheme, also seems obvious.
However, it was not actually followed for 2 of the 4 datasets for the CoNLL NER evaluations in 2002--3, as only the English and Dutch data were free of errors of this type (see Section \ref{sec:invalid-gold}).
As these datasets are arguably the most famous NER datasets in existence, this is surprising.
While this would only have a very minor impact on evaluation results, an evaluation cannot be reproducible if different scorers might interpret the gold standard differently due to differences in how invalid label sequences are handled (see Section~\ref{sec:scoring}).
When examining other NER datasets, we have found more pervasive occurrences of invalid label sequences.

We will not discuss guideline \ref{rule:stats} in any detail as practices change over time, but we will highlight the need to report a distribution of scores, rather than a single score.
\citet{reimers-gurevych-2017-reporting} demonstrate this clearly for NER specifically, and SeqScore supports aggregating scoring across multiple runs and reporting summary statistics.

Many of these rules may seem like common sense, but by enumerating them, we provide a published ``checklist'' for researchers to follow.

\section{The mechanics of NER evaluation}

We now turn to explaining the mechanics of NER evaluation to explain why following these guidelines is important.
In this section, we explain the subtleties of working with chunk encodings, which will reinforce the importance of following the first three guidelines.

\subsection{The CoNLL tradition}

Evaluating named entity recognition (NER) and similar chunking tasks is conceptually straightforward.
The primary metrics are the precision, recall, and F1 of the extracted chunks, often called phrases, or for NER specifically, entities or mentions.
The CoNLL-2000 shared task on chunking \citep{tjong-kim-sang-buchholz-2000-introduction} set the first and most long-lasting standard for distributing data for and evaluating chunking tasks.

Briefly, this standard---which we will call ``CoNLL-style"---is that each dataset (train, etc.) is represented in a sentence-split, tokenized, delimited format.
Each sentence consists of a sequence of lines, and each line contains at least a token and a label for that token.
This format was accompanied by a scoring script, \texttt{conlleval}.\footnote{\url{https://www.clips.uantwerpen.be/conll2000/chunking/conlleval.txt}}
The labels give information about the spans of the chunks, using encoding schemes that have developed from the original IOB representation of \citet{ramshaw-marcus-1995-text}.

While some models may use more complex encodings, the current standard for datasets is that chunks are encoded using BIO (begin, inside, outside), where the first token of each chunk receives a B- label, any following tokens in the chunk receive an I- label, and any tokens not contained in a chunk receive O label.
This standard format, albeit with minor variations, has been used continuously for NER datasets \citep[among others]{tjong-kim-sang-2002-introduction, tjong-kim-sang-de-meulder-2003-introduction, BenikovaBiemannKisselewetal2014, derczynski-etal-2017-results}.

Not every evaluation of these type of tasks has used this structure.
Many datasets \citep[e.g.,][]{doddington-etal-2004-automatic,hovy-etal-2006-ontonotes,strassel-tracey-2016-lorelei} use start and end offsets as the primary method of identifying spans, which can avoid issues related to tokenization and completely dissociates the annotation from the encoding of chunks using labels.
As we show later, this dissociation automatically removes a major source of non-reproducibility in evaluation.

\subsection{Scoring and repair}
\label{sec:scoring}

\begin{table*}[tb]
\small
\centering
\setlength{\tabcolsep}{10pt}
\begin{tabular}{l*{8}c}
\toprule
Encoding       & his & Liberal & Democratic & party & and & the & Russian & Duma \\
\midrule
Valid   & O  & B-ORG & I-ORG & I-ORG & O & O & B-MISC & B-ORG \\
\midrule
Invalid                   & O  & \textbf{I-ORG} & I-ORG & I-ORG & O & O & B-MISC & \textbf{I-ORG} \\ 
Begin Repair & O  & B-ORG & I-ORG & I-ORG & O & O & B-MISC & B-ORG \\
Discard Repair            & O  & O & O & O & O & O & B-MISC & O \\
\midrule
Stanza Repair           & O  & O & O & O & O & O & B-ORG & I-ORG \\
\bottomrule
\end{tabular}
    \caption{
    Valid and invalid BIO label sequences and repairs of the invalid sequence for the sentence fragment \emph{his Liberal Democratic party and the Russian Duma} from the CoNLL-03 English training data (lines 3633--40). 
    Labels that cause invalid transitions are bolded.
    }
    \label{tab:encoding}
\end{table*}

When it comes to evaluating system output, while the CoNLL-style format is truly simple, the process of using it for evaluation only \emph{seems} simple.

The fundamental problem is that there is generally nothing that forces a system's output---or even the annotation (see Section~\ref{sec:invalid-gold})---to follow the intended state machine of the scheme for encoding chunks.
While using a CRF may reduce---and constrained decoding \citep{lester-etal-2020-constrained} can eliminate---invalid label transitions, we must still be able to provide reproducible scoring methods for models that do not use these approaches.

As shown in Table~\ref{tab:encoding}, if we are using BIO encoding, a system could produce the sequence \texttt{O} \texttt{I-ORG}, illegally entering the ``inside'' state without going through ``begin.''
Similarly, we might encounter a \texttt{B-MISC I-ORG} transition, beginning a chunk of type MISC but then continuing into an ORG chunk.
Handling these invalid transitions requires an implicit or explicit \emph{repair} method.

\subsection{Repairs in practice}

Since the \texttt{conlleval} scoring script, scoring predicted labels and repairing the invalid sequences that they contain have gone hand in hand.
SeqScore follows this tradition, allowing for scoring labels that contain invalid sequences, but unlike \texttt{conlleval}, its repair methods are configurable, and unlike any other scorer we are aware of it supports inspecting the repaired label sequences through writing them to a file.
By requiring the user to select the repair method and making a previously invisible feature visible, we are making it easy for users to follow guideline \ref{rule:repair}.

The user can specify whether to perform \texttt{conlleval}-style repair, to \emph{discard} invalid sequences, or to make no repairs (\emph{none}), which will raise an error if any invalid sequences are encountered. 
The differences between these repair methods are show in Table~\ref{tab:encoding}.
Due to the complexities of attempting to repair invalid label sequences in  BIOES,\footnote{See \citet{kroutikov-f1-ner} for a discussion of the large number of ways to score invalid BIOES sequences.} repair is only supported for BIO and IOB encodings.

\begin{table}[tbh]
\small
\centering
\setlength{\tabcolsep}{4pt}
\begin{tabular}{l*{6}c}
\toprule
Repair & \multicolumn{5}{c}{Labels} \\
\cmidrule(r){1-1}
\cmidrule(l){2-7}
None & O & \textbf{I-ORG} & I-ORG & O & B-PER & I-PER \\
Begin & O & \textbf{B-ORG} & I-ORG & O & B-PER & I-PER \\
Discard & O & \textbf{O} & O & O & B-PER & I-PER \\
\bottomrule
\end{tabular}
    \caption{Original and repaired labels for \emph{Trade and Industry Secretary Ian Lang} (CoNLL-03 English)}
    \label{tab:o-i-repair}
\end{table}

\begin{table}[tbh]
\small
\centering
\setlength{\tabcolsep}{4pt}
\begin{tabular}{l*{5}c}
\toprule
Repair & \multicolumn{5}{c}{Labels} \\
\cmidrule(r){1-1}
\cmidrule(l){2-6}
None & O & B-ORG & I-ORG & \textbf{I-LOC} & O \\
Begin & O & B-ORG & I-ORG & \textbf{B-LOC} & O \\
Discard & O & B-ORG & I-ORG & \textbf{O} & O \\ 
\bottomrule
\end{tabular}
    \caption{Original and repaired labels for \emph{the Oceanic Control Center in} (CoNLL-03 English)}
    \label{tab:i-i-repair}
\end{table}

\begin{table}[tbh]
\small
\centering
\setlength{\tabcolsep}{4pt}
\begin{tabular}{l*{5}c}
\toprule
Repair & \multicolumn{5}{c}{Labels} \\
\cmidrule(r){1-1}
\cmidrule(l){2-6}
None & O & B-LOC & \textbf{I-ORG} & I-ORG & O \\
Begin & O & B-LOC & \textbf{B-ORG} & I-ORG & O \\
Discard & O & B-LOC & \textbf{O} & O & O \\ 
\bottomrule
\end{tabular}
    \caption{Original and repaired labels for \emph{( Rangoon ) University early} (CoNLL-03 English)}
    \label{tab:b-i-repair}
\end{table}

For example, given an I- followed by another I- of differing type such as \texttt{I-ORG} \texttt{I-LOC}, one could coerce either the first or second tag to match the other and maintain that this is all one mention. 
Another option is to treat the second tag as B- and begin a new mention.
The latter is what most scorers do, but it should be noted that this is not \emph{a priori} the correct choice.

As we describe each repair method in more detail, we will use examples from actual output on the CoNLL 2003 English data.
We used SeqScore to find the invalid transitions in the BERT \citep{devlin-etal-2019-bert} model output from \citet{tu-lignos-2021-tmr} and selected examples of each type.

For BIO, the possible invalid transitions are an I- preceded by O (Table~\ref{tab:o-i-repair}), an I- preceded by an I- of a different type (Table~\ref{tab:i-i-repair}), and an I- preceded by a B- of a different type (Table~\ref{tab:b-i-repair}).
\texttt{conlleval}-style scorers take the approach of changing any unexpected I- to a B- (thus our name ``begin repair''), while discard-style scorers discard tokens started by an invalid sequence.

While begin and discard are the dominant repair methods in use, other methods are possible. Stanza's \citep{qi-etal-2020-stanza} undocumented approach (shown in Table \ref{tab:encoding}) most closely resembles the discard repair method but does not discard all invalid sequences.
For invalid sequences caused by a type mismatch, Stanza uses the type of the last token as the type for the whole mention, and unlike begin or discard, keeps the entire span as a single mention.
For example, \texttt{B-ORG} \texttt{I-ORG} \texttt{E-LOC} would be decoded as one mention of type LOC, since LOC is the type of the last token. 
While we have described the repair methods that we are aware of, others may exist whether intentionally or as accidental deviations from more common repair methods. 

\citet{lignos-kamyab-2020-build} demonstrate the variation that occurs due to different repair methods for invalid label transitions, finding that at least one NER toolkit takes an alternate approach to handling invalid transitions that consistently produces higher F1 scores for some models than scoring with \texttt{conlleval}.
Its approach is not incorrect; these ``edge cases'' can be interpreted different ways.
However, the result is that different scorers can produce different scores for the same output, even though they claim to measure the same thing.

With these facts in mind, we believe that we are approaching a replicability crisis for NER and other chunking tasks, as scores cannot reliably be compared across papers, and replications can fail due to lack of information about the scoring procedure.

\subsection{Invalid transitions in gold standards}
\label{sec:invalid-gold}

Most discussions of invalid label sequences focus on the system output,
but widely-used annotated data often contains invalid sequences as well.
For example, the CoNLL-02 Spanish data is BIO-encoded but contains three invalid O to I- transitions, one in each of the train, testa, and testb subsets.
The original IOB-encoded CoNLL-03 German data contains 10 invalid transitions.\footnote{We validated the CoNLL-02 Dutch, CoNLL-03 English, GermEval 2014 \citep{BenikovaBiemannKisselewetal2014}, and W-NUT 2017 Emerging and Rare Entities \citep{derczynski-etal-2017-results} data sets and found no issues. The CoNLL-03 German data was corrected in a later BIO-encoded release.}

While they may not have major impacts on scores, these invalid sequences represent a replication issue.
Any scorer using the discard repair can \emph{remove mentions from the gold standard};
even if the number of mentions removed is small, it is not an acceptable evaluation practice for the scorer to effectively change the gold standard.
If two researchers use different repair methods and the annotation contains invalid transitions, they are not only evaluating their system output differently but also not evaluating against the same gold standard.

One of the design tenets of SeqScore is that the detection and repair of invalid label sequences is \emph{explicit} and \emph{configurable}.
SeqScore supports validating IO, IOB (IOB1), BIO (IOB2), and the isomorphic BIOES, BILOU, BMES, and BMEOW \citep{radford-etal-2015-named} encodings.

Here is an example of validating the CoNLL-02 Spanish training data using SeqScore:\\
{
\footnotesize
\texttt{
\$ seqscore validate --labels BIO esp.train\\
Encountered 1 errors in 1 tokens, 8323 sequences, and 1 documents in esp.train\\
Invalid transition O -> I-LOC for token 'San' on line 221619\\}
}

Our recommendation is that validation (and if needed, repair) be run on any invalid gold standards before scoring.
Doing so guarantees that the gold standard faithfully follows the chunk encoding and has no invalid transitions, so regardless of the repair method used, the gold standard will be interpreted the same way.
This practice satisfies guideline \ref{rule:gold}.

\subsection{Label conversion}

SeqScore also supports conversion between valid IO, IOB, BIO, BIOES, BMES, and BMEOW encodings using the \texttt{convert} subcommand.
To prevent malformed output, it raises an error if the input contains any invalid sequences.
The input can be repaired using the \texttt{repair} subcommand if it is IOB or BIO encoded.
By separating repair and conversion, there are no ``hidden'' changes.

Many other label scheme converters convert labels at the token (rather than mention) level, which allows invalid sequences to propagate from the input to the output, sometimes with unexpected results.
For example, Stanza converts to BIOES before scoring, passing along invalid label sequences from BIO to BIOES.
While in BIO invalid transitions are limited to invalid I- labels, when invalid BIO sequences are converted to BIOES, there are many potential ways to convert them depending on how the input was interpreted.


\section{Experiments}

Our paper so far has discussed the importance of following the proposed guidelines but has not quantified the impact of doing so.
We conducted a series of experiments on NER datasets to examine the extent to which the variations in scores from different repair methods applied to system outputs could lead to different results.
These experiments also demonstrate the usefulness of SeqScore as a package for producing a reproducible and complete set of results for sequence labeling tasks.

\begin{table}[tb]
\small
\centering
\begin{tabular}{l*{5}r}
\toprule
Lang. & Begin & Discard & $\Delta$ & p-value \\
\midrule
amh & 71.19 $\pm1.20$ & \textbf{71.87} $\pm1.11$ & 0.69 & 0.15 \\
hau & 89.78 $\pm0.41$ & \textbf{90.12} $\pm0.47$ & 0.34 & 0.19 \\
ibo & 84.18 $\pm0.94$ & \textbf{84.57} $\pm0.86$ & 0.39 & 0.33 \\
kin & 73.29 $\pm1.39$ & \textbf{74.51} $\pm1.26$ & 1.22 & 0.06 \\
lug & 80.02 $\pm0.90$ & \textbf{80.32} $\pm0.85$ & 0.30 & 0.29 \\
luo & 74.43 $\pm1.60$ & \textbf{74.96} $\pm1.56$ & 0.53 & 0.27 \\
pcm & 87.89 $\pm0.72$ & \textbf{88.48} $\pm0.71$ & 0.59 & \textbf{0.03} \\
swa & 87.43 $\pm0.55$ & \textbf{87.79} $\pm0.62$ & 0.36 & 0.17 \\
wol & 64.74 $\pm1.82$ & \textbf{65.19} $\pm1.70$ & 0.45 & 0.50 \\
yor & 77.63 $\pm0.17$ & \textbf{78.40} $\pm1.04$ & 0.77 & 0.07 \\
\bottomrule
\end{tabular}
    \caption{Comparison of F1 scores across repair methods using XLM-R and MasakhaNER data.}
    \label{tab:masakhane-xlmr}
\end{table}

\begin{table}[tb]
\small
\centering
\begin{tabular}{l*{5}r}
\toprule
Lang. & Begin & Discard & $\Delta$ & p-value \\
\midrule
hau & 86.87 $\pm0.38$ & \textbf{87.36} $\pm0.32$ & 0.49 & \textbf{0.01} \\
ibo & 84.82 $\pm0.77$ & \textbf{85.14} $\pm0.72$ & 0.32 & 0.32\\
kin & 72.14 $\pm1.07$ & \textbf{73.41} $\pm1.00$  & 1.27 & \textbf{0.02}\\
lug & 80.42 $\pm1.04$ & \textbf{80.83} $\pm1.05$  & 0.41 & 0.29\\
luo & 73.37 $\pm1.52$ & \textbf{74.18} $\pm1.53$ & 0.81 & 0.15\\
pcm & 87.97 $\pm0.62$ & \textbf{88.47} $\pm0.52$ & 0.50 & 0.10\\
swa & 86.73 $\pm0.49$ & \textbf{87.12} $\pm0.52$ & 0.39 & 0.13\\
wol & 65.35 $\pm1.58$ & \textbf{66.29} $\pm1.58$ & 0.94 & 0.26\\
yor & 78.96 $\pm0.86$ & \textbf{79.87} $\pm0.75$ & 0.91 & \textbf{0.03}\\
\bottomrule
\end{tabular}
    \caption{Comparison of F1 scores across repair methods using mBERT and MasakhaNER data.}
    \label{tab:masakhane-mbert}
\end{table}

As we show in the following experiments, NER using large multilingual models fine-tuned on lower-resourced datasets can show significant variation due to the scoring method used.
We selected lower-resourced datasets for two reasons.
First, we believe that this a major frontier for innovation in NER, and many new results will be reported in this area for years to come.
Second, unlike higher-resourced datasets, the current state of the art for these datasets involves the application of large language models, which are particularly prone to producing invalid transitions.

\subsection{Comparing repair methods}

We evaluated two multilingual models, multilingual BERT \citep{devlin-etal-2019-bert} and XLM-R \citep{conneau-etal-2020-unsupervised}, on the MasakhaNER datasets \citep{adelani21tacl} which cover 10 African languages. 
Both of the models were trained for 50 epochs, using 10 different random seeds; we report the mean and standard deviation (as mean $\pm$ std. deviation) of F1 across the seeds.
Amharic was excluded from the mBERT experiments, as mBERT was not trained on its character set and thus predicts no names. 
XLM-R was trained on all 10 MasakhaNER languages.

For each language, we report the difference in mean F1 score between the begin and discard repair methods for all languages except Amharic.
We also examine the difference in mean F1 score between two models that are scored using different repair methods.
We provide statistical significance of each comparison using the Wilcoxon rank-sum test, which is computed using the ten F1 scores for each configuration.
We use the Wilcoxon rank-sum test as it provides a robust comparison between two distributions without assuming that the scores are normally distributed, as there is no guarantee that scores follow such a distribution.


As shown in Tables \ref{tab:masakhane-xlmr} and \ref{tab:masakhane-mbert}, the discard repair method universally produces higher F1 scores.
Using the significance threshold of $p < 0.05$ (bolded), a handful of the comparisons between repair methods are statistically significant, specifically the mBERT scores for Hausa, Kinyarwanda, and Yoruba and the XLM-R scores for Nigerian Pidgin. 
The results that are not statistically significant still demonstrate a noteworthy difference in F1, with the discard score being 0.61 points higher than begin on average across all comparisons, statistically significant or not. 

It is not obvious why some models show a statistically significant difference with the discard repair method while others do not. 
Table \ref{tab:repair-counts} shows the average count of invalid label sequences across models and languages.
While it is notable that Kinyarwanda and Yoruba have higher counts and happen to be significant in experiments using mBERT, Hausa has a comparatively low count as does Nigerian Pidgin which also had significant results (see Tables \ref{tab:masakhane-xlmr} and \ref{tab:masakhane-mbert}). 

We also performed a small qualitative exploration of invalid transitions. 
We classified the invalid transitions in three ways: the begin strategy repairs in such a way that the repaired entity is correct, discard correctly discards a system predicted entity where there should be none, or neither correctly repairs the predicted entity.
In the case that neither repair method is correct, discard favors a higher F1 since the begin strategy creates a false negative and false positive, while discard creates only a false negative.

We examined the invalid transitions for XLM-R Nigerian Pidgin and Wolof---selected due to having the lowest and highest \emph{p}-values, respectively---in the test set output from the runs with the median scores.\footnote{As there were an even number of runs (10), we used the higher of the two median runs for each language.}
Wolof had 3 of 13 invalid transitions correctly repaired by begin, while discard correctly repaired only 2.
For Nigerian Pidgin, begin correctly repaired only 1 of 12 while discard correctly repaired 4 of 12.

While Nigerian Pidgin shows a larger gap between the effectiveness of the two repair methods, ultimately the number of repaired transitions is quite small due to their relative rarity and the small size of data sets for lower-resourced languages and thus it is difficult to draw conclusions.
Our analysis could not identify a simple explanation for the differences observed across languages, and this merits further examination in future work.

\subsection{Simulating a real scenario}

\begin{table}[tb]
\small
\centering
\begin{tabular}{l*{5}r}
\toprule
Lang. & XLM-R & mBERT \\
\midrule
amh & 13.9 $\pm4.33$ & - \\
hau & 11.5 $\pm3.27$ & 15.7 $\pm4.06$ \\
ibo & 19.8 $\pm6.64$ & 18.8 $\pm3.80$ \\
kin & 39.3 $\pm7.04$ & 40.3 $\pm8.87$ \\
lug & 14.9 $\pm3.92$ & 15.5 $\pm4.50$ \\
luo & 12.7 $\pm5.01$ & 14.9 $\pm3.93$ \\
pcm & 17.8 $\pm6.81$ & 12.9 $\pm5.59$ \\
swa & 15.0 $\pm3.23$ & 15.9 $\pm3.00$ \\
wol & 10.8 $\pm4.07$ & 17.3 $\pm5.36$ \\
yor & 32.4 $\pm6.92$ & 29.6 $\pm7.88$ \\
\bottomrule
\end{tabular}
    \caption{Means and standard deviations of across runs of the number of invalid transitions repaired for system output for each model and language.}
    \label{tab:repair-counts}
\end{table}

While we have shown that using a different repair method can sometimes lead to significant differences in F1 scores, using different repair methods on the exact same system output is not what happens in practice. 
Instead, the current situation is more likely to be that two different system outputs are unknowingly evaluated using differing repair methods by different authors.

\begin{table*}[tb]
\small
\centering
\begin{tabular}{l*{11}r}
\toprule
 & XLM-R & mBERT &  &  & & XLM-R & mBERT &  &  \\
Lang. & Begin & Discard & $\Delta$ & p-value & & Discard & Begin  & $\Delta$ & p-value \\
\midrule
hau & 89.78  & 87.36 & 2.42 & \textbf{0.0002} & & 90.12 & 86.87 & 3.25 & \textbf{0.0002} \\
ibo & 84.18  & 85.14 & 0.96 & \textbf{0.0233} & & 84.57 & 84.82 & 0.25 & 0.5708 \\
kin & 73.29  & 73.41 & 0.12 & 0.9397 & & 74.51 & 72.12 & 2.37 & \textbf{0.0012} \\
lug & 80.02  & 80.83 & 0.81 & \textbf{0.0494} & & 80.32 & 80.42 & 0.10 & 0.8206 \\
luo & 74.43  & 74.18 & 0.25 & 0.8206 & & 74.96 & 73.37 & 1.59 & \textbf{0.0257} \\
pcm & 87.89  & 88.47 & 0.58 & \textbf{0.0452} & & 88.48 & 87.97 & 0.51 & 0.1306 \\
swa & 87.43  & 87.12 & 0.31 & 0.2265 & & 87.79 & 86.73 & 1.06 & \textbf{0.0012} \\
wol & 64.74  & 66.29 & 1.55 & 0.0696 & & 65.19 & 65.35 & 0.16 & 0.8206 \\
yor & 77.63  & 79.87 & 2.42 & \textbf{0.0002} & & 78.40 & 78.96 & 0.56 & 0.3447 \\
\bottomrule
\end{tabular}
    \caption{Comparison crossing models and repair methods.}
    \label{tab:compare-model-repair}
\end{table*}
To simulate a more likely situation, suppose one researcher has trained an mBERT model and another has trained an XLM-R model but neither explicitly mentions what repair strategy was used while scoring. 
We will now explore how the use of different repair methods would affect the conclusions drawn from researchers unknowingly using different scoring procedures when comparing their models. 

In Table \ref{tab:compare-model-repair}, we compare XLM-R using the begin repair to mBERT using discard and XLM-R using the discard repair to mBERT using begin.
Suppose one team used XLM-R with the discard method to evaluate on Kinyarwanda while the other used mBERT and begin. 
The team using XLM-R with discard for scoring would have a score of 74.51 compared with the other team's score of 72.14, for a statistically significant difference in F1 of 2.37. 
If the teams switched scoring methods, the mean scores are much closer at 73.29 compared with 73.41, reducing the difference between the score to 0.12, which is statistically indistinguishable.

While Kinyarwanda is the most dramatic example, the difference in F1 changes considerably depending on the combination of repair methods used.
Of the 9 language datasets, 7 show a change in whether the difference between models is statistically significant depending on which repair method is used with each model, highlighting the important of guideline \ref{rule:repair}.
If in addition to using different repair methods, if one researcher reported their best test set score with no information about the distribution---as we have discovered a recent state of the art NER paper did---there would likely be even larger differences.

These experiments show that if researchers do not report their full scoring procedure, they may inadvertently obfuscate which models actually perform better and their claims of improvement may just be statistical noise.
However, if researchers use SeqScore, they can evaluate their system using multiple repair techniques without the risk of using two entirely different scorers, and their evaluations will be replicable by other researchers.

\section{Comparison with other toolkits}

To place SeqScore in the context of other work and to address the question of novelty, we compare SeqScore to other similar tools.

\begin{table*}[tbh]
\centering
\setlength{\tabcolsep}{3pt}
\footnotesize
\begin{tabular}{l*{9}c}
    \toprule
    \textbf{} & \textbf{SeqScore} & \textbf{Stanza} & \textbf{NCRF++} & \textbf{iobes} & \textbf{sighsmile} & \textbf{spyysalo} & \textbf{wnuteval} & \textbf{seqeval} & \textbf{conlleval} \\
    
    \midrule
    Warns for invalid label seqs. & \cmark & & & \cmark & & & \cmark & &  \\
    begin repair & \cmark & & & & \cmark & \cmark & \cmark & \cmark & \cmark\\
    Discard repair & \cmark & \cmark & \cmark & \cmark & & & & \cmark & \\
    Converts label schemes & \cmark & \cmark & \cmark & \cmark & & & & & \\
    Scoring & \cmark & \cmark & \cmark & & \cmark & \cmark & \cmark & \cmark & \cmark\\
    Aggregation across runs & \cmark & & & & & & & & \\
    \bottomrule
\end{tabular}
  \caption{Comparison of package features}
  \label{tab:comaprisons}
\end{table*}

\subsection{Design Considerations}

The primary goal of SeqScore is to provide a highly usable scorer for chunk extraction sequence labeling tasks such as NER.
But SeqScore is not just a scorer; it is designed to address the entire lifecycle of working with data: validating and examining annotation, converting between various chunk encodings, identifying and repairing invalid label sequences, and finally producing scores.
SeqScore is implemented in Python, and like Git it uses subcommands to perform each task, for example \texttt{score} to score, and \texttt{validate} to validate files.
While other packages exist for scoring and handling invalid label sequences, no other package has the convenience of everything in one place.

We believe this convenience lessens the barriers to providing more detailed reporting of scoring methods, and that this convenience and packaging together of all of these features is a novelty of SeqScore.
For example, unlike \texttt{conlleval} and every other NER scorer we examined, SeqScore supports aggregating scores across multiple prediction files for the same reference.
This enables the now-common practice of reporting the mean and standard deviation across runs, aiding in following  guideline \ref{rule:stats}.
While this is a simple feature, by reducing the effort required to report these scores, we believe we can help improve adoption of this practice.
Of the papers we surveyed, only about two-thirds were clear about how many runs they used and whether their reported score was an average or a best run.

Table~\ref{tab:comaprisons} compares the features of SeqScore against other packages for scoring and working with sequence labeling data for chunking tasks. 
While SeqScore is designed to include as many features as possible, there are some features it does not implement.
One is partial match scoring, which is implemented in nervaluate \citep{segura-bedmar-etal-2013-semeval}\footnote{\url{https://github.com/ivyleavedtoadflax/nervaluate}} following the MUC scoring approach \citep{chinchor-sundheim-1993-muc}.
Also, SeqScore only processes CoNLL-style file formats.

NER scorers can broadly be grouped in terms of how closely they resemble the \texttt{conlleval} Perl script. 
There are direct re-implementations, those that score in the same spirit as \texttt{conlleval} but have additional features, and those that take a different approach to invalid labels.
The set of scorers we examine is not exhaustive but covers the most widely-used ones.


\subsection{\texttt{conlleval} reimplementations}
To the best of our knowledge after testing on a number of datasets and edge cases, each of these is a faithful replication of the original \texttt{conlleval} Perl script: spyysalo \texttt{conlleval.py}\footnote{\url{https://github.com/spyysalo/conlleval.py}}, and sighsmile \texttt{conlleval.py}\footnote{\url{https://github.com/sighsmile/conlleval}}.
The spyysalo and sighsmile re-implimentations differ from the original \texttt{conlleval} script mainly in that they have support for BIOES and are written in Python instead of Perl.

\subsection{\texttt{conlleval}-style scorers}

\paragraph{\texttt{wnuteval}\footnote{\url{http://noisy-text.github.io/2017/files/wnuteval.py}}}
 \texttt{wnuteval} is limited to the entity types used in the shared task it was developed for.
 It raises warnings about invalid transitions.
 However, it does not handle multiple mention encoding schemes.
 We also found that it does not raise any warnings or errors about uneven document lengths between system and gold files; while this seems like an unusual case to test, with the use of models set maximum sequence lengths as a decoding hyperparameter, it is common to accidentally truncate sentences when producing system output for evaluation.

 \paragraph{seqeval}
 seqeval \citep{seqeval} can score on many different label schemes. 
 It is unique in being one of the only scorers we examined that has more than one approach to invalid label sequences, a feature added concurrently with the development of SeqScore.
 Seqeval refers to them as \emph{default} and \emph{strict} modes. \emph{Default} is \texttt{conlleval}-style (begin), while \emph{strict} is what we refer to as \emph{discard}.

\subsection{Internal scorers}
There are also numerous internal scorers that are part of larger NLP toolkits and packages. 
Though there are certainly plenty of others, we examine evaluation methods found in NCRF++ and Stanza. 
Neither of the approaches of these internal scorers follow \texttt{conlleval}-style handling of invalid label sequences.
\paragraph{Stanza}
Stanza \citep{qi-etal-2020-stanza} is a collection of models and tools for NLP. 
It supports NER in multiple languages and includes its own scorer implementation. 
Stanza's scorer is similar to \emph{discard} or seqeval's strict mode, with a few exceptions. 
Stanza also includes a number of tools for converting between schemes.  
\paragraph{NCRF++}
NCRF++ \citep{yang-etal-2018-design} is a framework for doing neural sequence labeling tasks in a highly configurable way. 
It implements its own scorer with an approach to invalid sequences using the \emph{discard} method. 

\subsection{Handling invalid label transitions}

\citet{lester-2020-iobes} provides a library for parsing label schemes, identifying invalid label sequences, converting between label schemes, and enumerating the legality of possible transitions. While this library is very useful for handling label schemes and invalid transitions, it does not address what we believe is a necessary decoupling of the scorer from the handling of invalid label sequences. While it is capable of identifying invalid transitions and supporting one’s own implementation to constrain or repair invalid sequences, it does not provide common methods for repairing invalid sequences. 

 \citet{lignos-kamyab-2020-build} demonstrate the difference that can occur when two scorers handle invalid label sequences differently.
 However, they do not provide any software to evaluate these differences and only test using CoNLL-03 English data with older neural models.

\section{Conclusion}

This paper has provided guidelines for reproducible NER research and demonstrated the importance of following them, both by describing the principles behind them and demonstrating the impact on actual scores.
Ultimately, researchers can choose to either accept the status quo---which for NER is non-reproducible research due to both a lack of standard practices and a lack of standard tools---or attempt to elevate the practice in the field to higher standards.
We hope that by providing a software toolkit to help follow these guidelines, we have substantially reduced the barriers to performing reproducible research for this task.

While we have created a software package to accompany our recommendations---one that makes following them extremely simple---we do not claim that using SeqScore is necessary for reproducible results.
Just like sacreBLEU is not the only way to produce a reproducible MT score, SeqScore is not the only reproducible way to score NER output.
However, as it is actively maintained, well-tested, and it can handle multiple repair methods, we strongly encourage its use.
By focusing on transparency and prioritizing supporting reproducible research in the design of SeqScore, we believe we have produced a toolkit that can have substantial positive impact on the field.

Adoption of this paper's recommendations by researchers will increase transparency in the scoring process and enable standardization of scoring methods in a field we believe is approaching a reproducibility crisis.

\section*{Acknowledgments}
We would like to thank David Adelani for helping us work with the MasakhaNER data. 
We would also like to thank two anonymous reviewers for their useful feedback on the paper. 
Chester Palen-Michel was supported by the Alfred Schonwalter Graduate Computer Science Summer Fellowship through a gift from a generous donor to Brandeis University.

\bibliographystyle{acl_natbib}
\bibliography{anthology,additional}


\end{document}